\documentclass[sigconf,natbib=true,anonymous=false]{acmart}

\AtBeginDocument{%
  \providecommand\BibTeX{{%
    \normalfont B\kern-0.5em{\scshape i\kern-0.25em b}\kern-0.8em\TeX}}}

\copyrightyear{2021}
\acmYear{2021}
\setcopyright{acmcopyright}\acmConference[CIKM '21]{Proceedings of the 30th ACM International Conference on Information and Knowledge Management}{November 1--5, 2021}{Virtual Event, QLD, Australia}
\acmBooktitle{Proceedings of the 30th ACM International Conference on Information and Knowledge Management (CIKM '21), November 1--5, 2021, Virtual Event, QLD, Australia}
\acmPrice{15.00}
\acmDOI{10.1145/3459637.3482405}
\acmISBN{978-1-4503-8446-9/21/11}

\acmSubmissionID{rgfp1565}

\definecolor{blue(pigment)}{rgb}{0.2, 0.2, 0.6}

\newtheorem{definition}{\textbf{Definition}}

\newcommand{\remove}[1]{}

\newcommand{\xhdr}[1]{\vspace{1mm}\noindent{{\bf #1.}}}

\newcommand{\cnote}[1]{
\textcolor{violet}{$\ll$\textsf{#1}$\gg$}
}

\newcommand{\ingredients}{O_s}
\newcommand{\instructions}{I_s}
\newcommand{\nstart}{\textit{START}}
\newcommand{\nend}{\textit{END}}

\usepackage{paralist}

\begin{document}
\fancyhead{}

\title{50 Ways to Bake a Cookie: Mapping the Landscape of Procedural Texts}

\author{Moran Mizrahi}
\affiliation{%
\institution{The Hebrew University of Jerusalem}
\city{Jerusalem}
\country{Israel}}
\email{moranmiz@cs.huji.ac.il}

\author{Dafna Shahaf}
\affiliation{%
\institution{The Hebrew University of Jerusalem}
\city{Jerusalem}
\country{Israel}}
\email{dshahaf@cs.huji.ac.il}

\begin{abstract}
 The web is full of guidance on a wide variety of tasks, from changing the oil in your car to baking an apple pie. 
 However, as content is created independently, a single task could have {thousands} of corresponding procedural texts. 
 This makes it difficult for users to view the bigger picture and understand the multiple ways the task could be accomplished. 
 In this work we propose an unsupervised learning approach for {\bf summarizing multiple procedural texts} into an intuitive graph representation, allowing users to easily explore commonalities and differences. We demonstrate our approach on recipes, a prominent example of procedural texts. 
 User studies show that our representation is intuitive {and coherent} and {that it} has the potential to help users with several sensemaking tasks, including adapting recipes for a novice cook and finding creative ways to spice up a dish.
\end{abstract}






\maketitle

\section{Introduction} \label{introduction}


Procedural texts play an important part in our lives: recipes, how-to instructions, scientific procedures, navigating directions and manuals are only a few common examples. The web includes procedural texts on a variety of topics, from recipe websites\footnote{\href{www.allrecipes.com}{\color{blue(pigment)} allrecipes.com}, \href{www.epicurious.com}{\color{blue(pigment)} epicurious.com}} to Maker websites\footnote{\href{https://www.instructables.com/}{\color{blue(pigment)} https://www.instructables.com}} and general how-to websites\footnote{\href{www.Wikihow.com}{\color{blue(pigment)} Wikihow.com}, \href{www.eHow.com}{\color{blue(pigment)} eHow.com}}.


\begin{figure*}[t!]
\includegraphics[width=\textwidth]{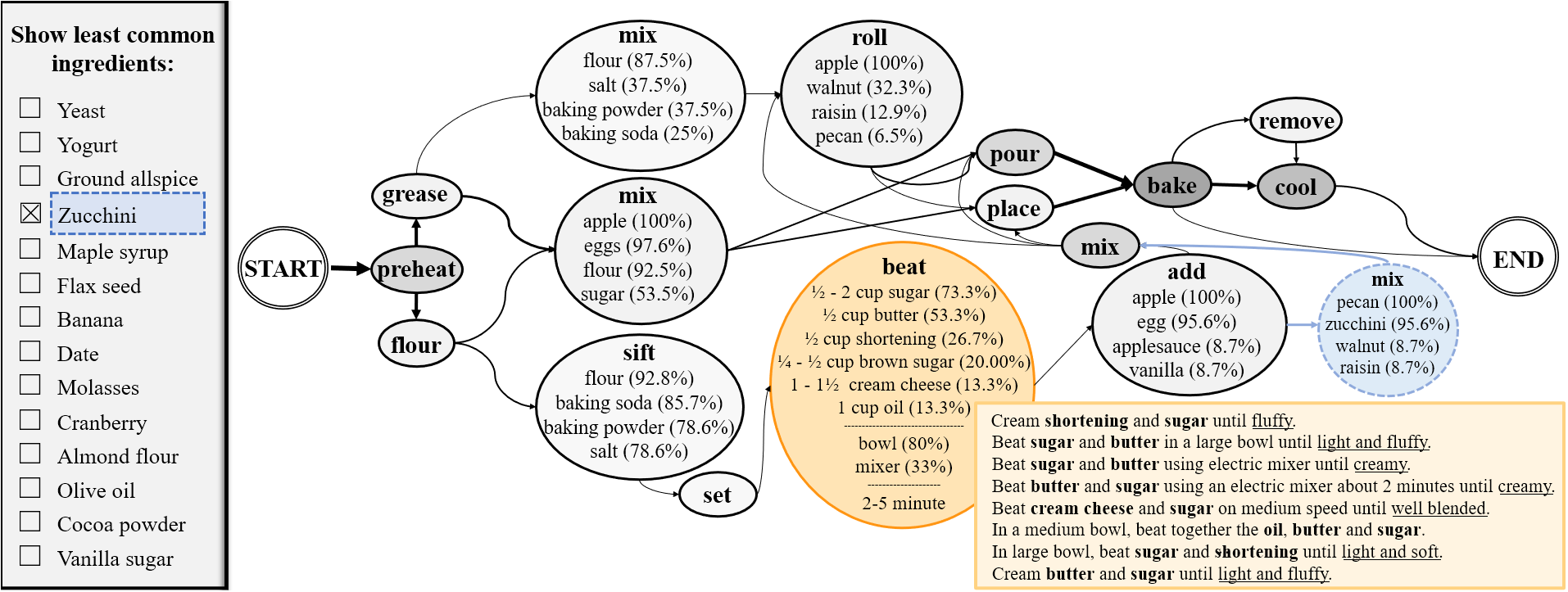}
\caption{\label{fig:main} 
Summary graph for apple cake recipes. Each node represents a cluster of similar instructions. Darker nodes indicate larger clusters; thicker edges indicate strong connections. Paths correspond to execution plans. Nodes show a compressed summary of their instructions: main cluster verb and its most frequent ingredients. The ``beat'' node is clicked on, showing a full summary, including ingredient quantity ranges, cooking instruments and preparation time range. Next to the ``beat'' node is a sample of its associated instructions. On the left  is a list of the least common ingredients. Clicking on an  ingredient that does not appear in the graph reveals hidden paths with this ingredient (in blue). 
}
\end{figure*}

However, a single task might have
thousands of corresponding procedural texts.
This is both due to variations (for example, different recipes for the same dish) and due to the distributed nature of the web, where content is created independently by people who do not communicate.
Thus, looking at one (or a few) procedural texts only gives the reader a limited view of the possibilities. 
Consequently, when people try to determine the best choice for them given preferences (e.g., taste) and constraints (budget, time, items they do or do not have), they often resort to extensive browsing and comparisons between different texts to get the bigger picture.

Automatic understanding of procedural texts is a difficult problem, requiring capturing the interplay between entities, attributes and their dynamic transitions. There has been a recent surge in work in understanding procedural texts \cite{mishra2018tracking,bosselut2017simulating,gupta2019tracking,du2019consistent,tandon2018reasoning,amini2020procedural} 
and in visualizing procedural texts as graphs \cite{karikome2018flow,chen2011learning,patow2010user,yamakata2016method,kiddon2015mise,mori2014flow,maeta2015framework}. However, these works focus exclusively on analyzing a \emph{single} procedural text. 
In contrast, our goal in this paper is to automatically {\bf summarize and organize many texts sharing the same goal}, allowing users to explore \emph{commonalities and differences} between texts at a glance. We envision a system that will guide the user in finding a way to complete the task that best fits their \emph{preferences or constraints}. Importantly, the chosen alternative could be a \emph{modification} or \emph{combination} of the original texts.

We focus on recipes, a prominent example of procedural texts. We build a system taking as input multiple recipes for the same dish. The output is an intuitive graph representation, mapping the entire landscape of variations. See Figure \ref{fig:main} for a summary of $\sim200$ apple-cake recipes. A node in the graph corresponds to a set of similar actions, and a directed path represents a way to make the dish. 
Alternative paths indicate different approaches, such as creaming butter and sugar before adding the other ingredients (bottom path) or mixing them all in a single step (middle path). The graph makes it easy to identify common ingredients (flour, apples, baking powder, sugar and eggs) and techniques, as well as  anomalies that could potentially spark innovative ideas, such as using yogurt, allspice or zucchini, or using a microwave to bake the cake instead of an oven. 

We believe such a representation could be especially useful to users who want to adjust a recipe to meet specific preferences or needs; to novice cooks looking to avoid rookie mistakes (e.g., many recipes do not explicitly mention the need to rinse grains or let the meat rest); and to anyone looking for new ideas for spicing up a familiar dish. Our contributions are:




\begin{itemize}
\item We propose a novel approach for summarizing procedural texts {sharing a goal} into an intuitive graph representation. 

\item  We demonstrate our approach on cooking recipes. We devise an unsupervised recipe parser, taking into account the unique structure of recipes. {We believe the principles behind the parser could be generalized to other domain-specific parsers for procedural texts.} We then propose a general-purpose algorithm for constructing the summarization graph. 

\item We assess the quality of our pipeline's individual components and conduct a user study to evaluate our representation in terms of \emph{intuitiveness} (can users understand it with no explanation?), \emph{coherence} (do paths correspond to recipes?) and \emph{utility} (can it help users performing sensemaking\footnote{Sensemaking \cite{russell1993cost} is the task of constructing a mental representation of interrelated pieces of information, often in the context of 
understanding large document collections.} tasks?). 
User studies demonstrate that our representation is intuitive and coherent. 
Evaluation by cooking experts shows the graph users perform better than users of a baseline interface at several sensemaking tasks, including adapting recipes for novice users and finding creative ingredients. 
 \item We release open-source code 
 at \href{https://github.com/moranmiz/50-Ways-to-Bake-a-Cookie}{\color{blue(pigment)} https://github.com/moranmiz}
 \href{https://github.com/moranmiz/50-Ways-to-Bake-a-Cookie}{\color{blue(pigment)} /50-Ways-to-Bake-a-Cookie}.
\end{itemize}

We believe that our representation could serve as foundation for future systems that digest a large set of procedural texts. We are particularly excited by the potential of such system to \emph{synthesize} new procedures, rather than simply recommend existing ones.




\begin{figure*}
\centering
\includegraphics[width=1\textwidth]{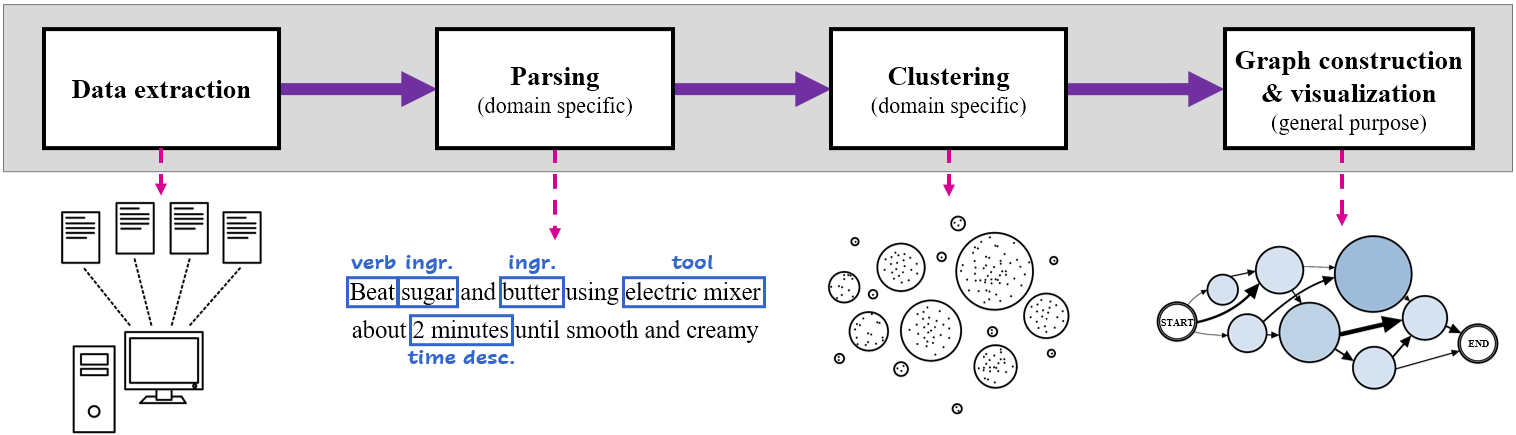}
\caption{\label{fig:scheme} 
A scheme demonstrating our general approach to constructing the summarization graph. The stages are: gathering data, parsing it, clustering instructions based on predefined similarity measure, constructing the graph and visualizing it.
}
\end{figure*}

\section{Problem definition} \label{task_definition}

Given a large set of procedural texts sharing the same goal, we wish to summarize these texts in a way that will help users view the big picture. In particular, we want to find a representation that will (1) allow the user to explore {\bf commonalities and differences} between the ways to complete the task, (2) make it easy for the user to choose a way to complete the task, satisfying personal {\bf preferences or constraints}. Importantly, the chosen way need not be one of the original procedural texts, but rather could be a modification (or combination) of the original texts. More formally,


\begin{definition}[\textbf{Summary Graph}] \label{def:def1}
Let $\mathcal{S}$ be a set of procedural texts sharing the same goal. Each $s\in\mathcal{S}$ is a pair $\left(\ingredients,\instructions \right)$,
where $\instructions$ is a sequence of instructions, and $\ingredients$ is a set of objects needed to carry out the instructions. 
Our goal is to construct a \emph{summary graph} $G_{\mathcal{S}}=(V,E)$. Each \emph{node} in $V$ represents a set of (semantically similar) instructions from $\mathcal{S}$.
There are also two special nodes, \nstart\ and \nend. Directed paths from \nstart\ to \nend\ represent ways toward achieving the goal.
Nodes are weighted and labeled; edges are weighted.
\end{definition}

\remove{
\begin{definition}[\textbf{Node Summary}] \label{def:def2}
\cnote{Let $v$ be a node in $G_{\mathcal{S}}$ that represents a set of (semantically similar) instructions. Our goal is to summarize its content. A \emph{node summary} contains the action that represent the set along with a statistical summary of the objects needed to carry out the instructions and execution time-range.}
\end{definition}
}




Figure \ref{fig:main} shows an example for summary graph, summarizing $\sim200$ apple-cake recipes.
To facilitate exploration, we provide several visual cues: dark nodes contain more instructions, and thick edges represent strong connections between the nodes. Nodes could contain hundreds of instructions, and thus we need to summarize their contents for the visualization. 
For recipes, we specify the action (e.g., ``mix'', ``bake'') along with a statistical summary of the ingredients, tools, and execution time-range. 
Left click on a node reveals quantities (see ``beat’’ node); right click shows its actual (natural-language) instructions (see a sample near ``beat''). 


The graph representation gives a general overview that allows users to explore different ways to bake a cake
and better understand the process.
For example, consider the highlighted ``beat’’ node. 
Looking at this node, one could deduce that 
butter, appearing in $53.3\%$ of the instructions, is more popular than shortening or oil, that this step could use a mixer and only takes a few minutes. 
%
The thick edge from node ``bake'' to node ``cool'' indicates that cooling the cake after baking it is crucial.
Alternative paths indicate different approaches, such as creaming butter and sugar before adding the other ingredients (bottom path) or mixing them all in a single step (middle path).

The graph interface also makes it easy to identify anomalies that could potentially spark innovative ideas. 
Expanding the ``bake'' node, we observe it is possible to bake a cake using a microwave.\footnote{\href{https://tinyurl.com/apple-mug-cake}{\color{blue(pigment)} https://tinyurl.com/apple-mug-cake}} The rare ingredient list includes ingredients such as allspice and even zucchini (interestingly, the rarest ingredient is yeast; upon further examination, we realized that the vast majority of cake recipes  are indeed risen by baking soda/powder\footnote{\href{https://tinyurl.com/yeast-leavened-cake}{\color{blue(pigment)} https://tinyurl.com/yeast-leavened-cake}}).

\section{Implementation} \label{implementation}
%

Before diving into the details, we give a general overview of our approach towards constructing the summary graph, illustrated in Figure \ref{fig:scheme}.
First, we gather {data} of procedural texts sharing the same goal. 
Second, we {parse} the data {using our unsupervised parser. Then, we take advantage of the structure extracted by our parser to} 
define a similarity measure between instructions and {cluster} similar instructions. These clusters  constitute the graph's {nodes}. Next, we connect nodes so that every {path} corresponds to an execution plan. As the resulting graph might be noisy and too large to visualize effectively, we {prune} it, reducing noise in the process. Note that 
while the first three steps in the scheme (gathering data, parsing, similarity) are task-dependent, the final step is general. Code is available at \href{https://github.com/moranmiz/50-Ways-to-Bake-a-Cookie}{\color{blue(pigment)} https://github.com/moranmiz/50-Ways-to-Bake-a-Cookie}.

\section{Data collection} \label{data}
{We gather data of procedural texts sharing the same goal.
For recipes, we crawled Allrecipes.com for 18,976 recipes of 98 popular dishes.}
%
{The average number of ingredients per recipe is 10.11 (std=4.09). The average number of instructions per recipe is 3.86 (std=1.865) before tokenization, and 12.65 (std=6.65) after tokenization (see Section \ref{parser}). 
The average number of words per recipe is 162.01 (std=76.13). Considering instructions only, the average number of words is 116.33 (std=64.7).
The vocabulary size is 9322.}

{In Section \ref{word2vec_sec}, we construct a word2vec model on recipe instructions. For this step, we also use the instructions of 97,862 recipes from "Now You're Cooking!"\footnote{Data is available at  \href{http://www.ffts.com/recipes.htm}{\color{blue(pigment)} http://www.ffts.com/recipes.htm}.\label{now_youre_cooking}}. The average length of a recipe in this dataset (considering instructions only) is  63.45 words (std=46.13). The vocabulary size of this additional dataset is 44,601.}

\section{Model}

\subsection{Unsupervised parser}\label{parser}
Referring back to Definition~\ref{def:def1}, {in our use case} $\mathcal{S}$ is a set of \textit{cooking recipes for the same dish}. Each recipe $s\in \mathcal{S}$ is a pair $\left(\ingredients,\instructions \right)$, where $\ingredients$ is a set of \textit{ingredient objects}, and $\instructions$ is a \textit{sequence of instructions}. We define an ingredient object $o\in \ingredients$ as a tuple consisting of the ingredient’s quantity, quantity unit and name. An instruction object $i\in \instructions$ consists of the instruction's main verb (e.g. ``mix'' for ``mix all the ingredients''), sets of ingredient objects and instrument names that appear in the instruction, and an instruction's time range tuple (indicating minimal and maximal duration). 

{We want to parse natural-language recipes into our representation
and use the structure to compare instructions.}
We have experimented with off-the-shelf parsers, including open-IE \cite{stanovsky2018supervised} and UDPipe \cite{straka2017tokenizing}.
However, recipe data has several unique characteristics and challenges, and thus we decided to implement our own  parser.





One prominent challenge is that in recipes, the same ingredient is often referred to in multiple forms. For example, the ingredient list might mention specific ingredients as ``ground nutmeg'' and ``cinnamon'', but the instructions will refer to ``spices'' (\emph{generalization}). Similarly, the ingredient list might mention ``Granny Smith apples'', but the instructions will only mention ``apples''.
We refer to that specific type of generalization as \emph{abbreviation}.






Keeping track of abbreviations and generalizations has two important advantages: 
when parsing instructions such as ``sift sugar and spices'', we can identify the implicit list of ingredients. 
When comparing different recipes, we can better measure similarity between ingredients. For example, we can conclude that “vanilla” is similar to “vanilla extract”, but “bread” and “bread crumbs” are certainly different.

Thus, our parser implements two complementary tasks -- ingredient parsing and instruction parsing. We found it helpful to take into account the \emph{full instruction text} when parsing ingredient lines, and the ingredient list when parsing instruction lines.

\xhdr{A note on generalization}
Abbreviations and generalizations are common in many procedural texts, from material science to make and craft instructions. Hence, we believe similar methods could be helpful when implementing other domain-specific parsers as well.

\begin{figure}[t!] \label{parser_img}
\centering
\includegraphics[width=1.00\linewidth]{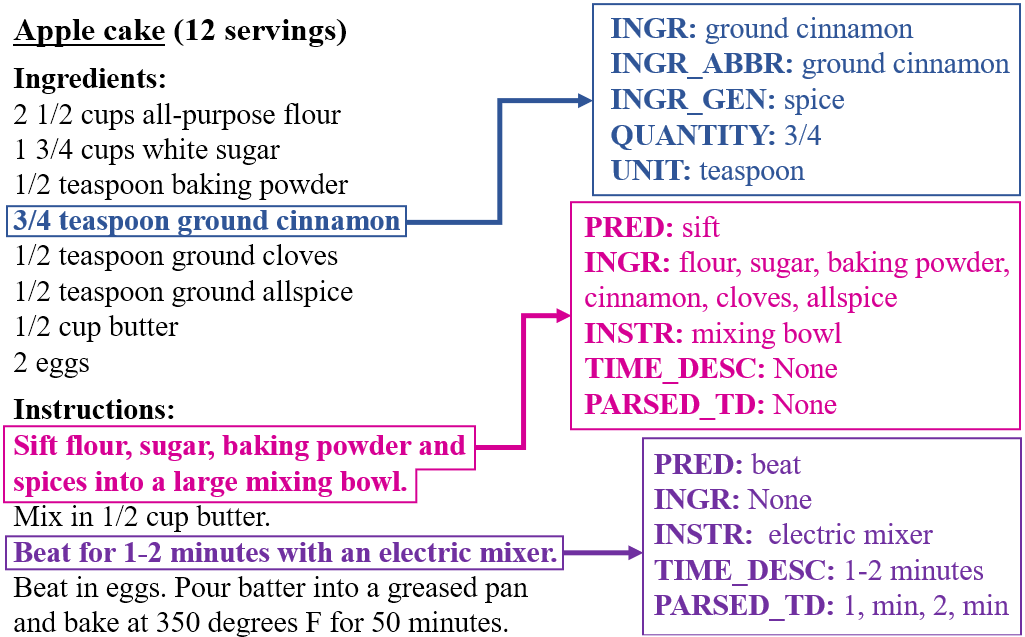}
\caption{\label{fig:parser} 
Parser outputs for an apple-cake recipe. The upper rectangle is an \emph{ingredient} parsing output in which ``ground cinnamon'' is the parsed ingredient and ``spice'' is its generalization. No abbreviation for ``ground cinnamon'' was explicitly found in text and thus the abbreviation is identical to the parsed ingredient; 
The two rectangles below are \emph{instruction} parsing outputs. In the upper one, the parser managed to extract ``cinnamon'', ``clove'' and ``allspice'' from ``spices''.
}
\end{figure}

\xhdr{Ingredient parsing} 
Our ingredient-line parser first parses the ingredient lines and tries to extract the ingredient name, quantity and unit using regular expressions. 
In addition, the parser also looks for abbreviations and generalizations in the text.

To derive an ingredient \emph{abbreviation} we lemmatize the instruction text and search for the longest consecutive word sequence that the ingredient name shares with the text. If there are several longest sequences, we prefer one that ends with a noun (an ingredient's abbreviation is usually consecutive adjectives followed by a noun). 


A failure to find an abbreviation is usually caused by a more generalized description in the instructions
(e.g., ``spices'' for ``ground cinnamon'').
To derive
a \emph{generalization} of a missing ingredient,
we remove the already-found abbreviations. Then, for every noun in the text (e.g., ``spice''), we  use  WordNet \cite{fellbaum2012wordnet}  and check whether ``food'' or ``fruit'' is one of its hypernyms. 
In this case, if the noun is also a hypernym of the missing ingredient, we consider it the ingredient's generalization {(see example outputs in Figure \ref{fig:parser}).}

{We note that 
the ratio of recipes containing a generalization in our data is 8.56\%, and that this ratio varies significantly among different dishes (e.g., close to 0\% for ``deviled eggs'' and around 38\% for ``whole-grain bread'').}


\xhdr{Instruction parsing}
When parsing an instruction, we want to extract its main verb, ingredients, tools
and preparation time range. 
%
For obtaining tools and preparation time we rely on regular expressions.
To derive the main verb we build upon the coreNLP parser \cite{Manning14thestanford}.
Applied to the raw data, the coreNLP parser finds the correct verb for only $\sim75\%$ of the sentences, perhaps due to the imperative form (which is rare in training data). Thus, we concatenate the prefix ``You should'' to the instructions. If {still} no verb is found by the parser, we look up verbs from a list of common cooking verbs {(the collection of all the verbs we managed to parse before)}. 
{As noted above, identifying the ingredients is done using the extracted abbreviations and generalizations.}
Refer to Figure \ref{fig:parser} for output examples. 


\remove{
To derive an ingredient \emph{abbreviation} we lemmatize the instruction text and search for the longest consecutive word sequence that the ingredient name shares with the text. If there are several longest sequences, we prefer one that ends with a noun (an ingredient's abbreviation is usually consecutive adjectives followed by a noun). 


A failure to find an abbreviation is usually caused by using a generalized description in the instructions
(e.g., ``spices'' to describe ``cinnamon'').
Thus, to derive
a \emph{generalization} of a missing ingredient,
we remove the already-found abbreviations. Then, for every noun in the text (e.g., ``spice''), we  use  WordNet \cite{fellbaum2012wordnet}  and check whether ``food'' or ``fruit'' is one of its hypernyms. 
If the noun is indeed a hypernym of the missing ingredient, we consider it the ingredient's generalization.
}


%
%


\remove{
\subsubsection{\textbf{Instruction parsing}}
When parsing an instruction, we want to extract its main verb, ingredients, tools
and preparation time range. 
%

We use the coreNLP parser \cite{Manning14thestanford} to derive the main verb and its arguments. Applied to the raw data, the parser finds correct arguments for only $\sim75\%$ of the sentences, perhaps due to the imperative form (which is rare in training data). Thus, we concatenate the prefix ``You should'' to the instructions. We rely on regular expressions for obtaining tools and preparation time. If no verb is found by the parser we look up verb from a list of common cooking verbs. Figure \ref{fig:parser} shows some output examples.

We manually evaluated our parser on 100 ingredient lines and 100 instruction lines, randomly selected. The classifier achieved $94\%$ accuracy on ingredient lines, $92\%$ accuracy on instructions, and $97\%$ accuracy identifying the main verbs. 
}

%

%

\xhdr{Sentence tokenization}
{In the recipes of our dataset, one line in the instructions often corresponds to multiple actions.
We divide these instructions into sub-instructions that are as concise and as simple as possible. To do so, we first tokenize the raw instructions using the  coreNLP sentence tokenizer.
Then, we break down complex instructions consisting of several verbs, using the common cooking verbs list  found by our parser. 
For example, the instruction: ``Combine the water, 1/2 cup sugar and chocolate in a saucepan and cook over low heat just until the chocolate melts'' is divided into: ``Combine the water, 1/2 cup sugar, and chocolate in a saucepan'' and ``cook over low heat just until the chocolate melts''. 
This {last} step {(breaking down complex instructions)} affects 14.43\% of the instructions in the data. 
}

\xhdr{Evaluation}
We manually evaluated our {unsupervised} parser on {200} ingredient lines and {200} instruction lines, randomly selected. 
{For ingredients, the parser achieved accuracy of 93.5\% for extracting the ingredients, 95.5\% and 97\% for deriving abbreviations and generalizations, and 100\% and 99.5\% for parsing amounts and units.}
{As for the instructions, our parser succeeded in extracting the right verb for 93.5\% of them, the instrument for 95.5\% of them and the time description for 100\%.
Moreover, it identified correctly 95.26\% of the ingredients appearing in them. 

In comparison, open-IE \cite{stanovsky2018supervised}
identified the right verb for only 76.5\% of the instructions (failing to extract anything for 17\% of them). It was also very difficult to infer ingredients or tools from the output (Representative outputs: [V: Mix] [ARGM-LOC: in onion , cilantro , tomatoes] [ARG1: , and garlic] , [V: shortening][ARG1: Cream] -- for ``Cream shortening'', [V: Add] [ARG1: the sugar and vanilla and beat well]). 
On the other hand, UDPipe \cite{straka2017tokenizing} 
found the right verb for 82.5\% of the instructions. Failures are often due to identifying verbs as nouns (Representative outputs: [N: Cover][N: skillet], [N: Spoon][N: mixture][ADP: into][N: cups]).

We note a recent relevant work by Diwan et al. \shortcite{diwan2020named}, suggesting a NER model to infer recipe instruction structure. We could not compare our results to theirs as the authors released only partial code and data at the time of completing this paper.
}

\subsection{Clustering} \label{similarity_measure}
\label{clustering} 

As noted earlier, each node in the graph corresponds to a set of semantically similar instructions. 
Similar instructions could be, for example, ``cream shortening and sugar until fluffy’’ and ``beat butter and sugar using an electric mixer about 2 minutes until creamy’’ as shown in Figure \ref{fig:main}. 
%

It is not straightforward to measure how semantically close two cooking instructions are. 
For example, consider the following instructions (taken from apple-cake recipes): 
\begin{enumerate}
\item ``Toss together the shredded apple, cinnamon and sugar in a bowl until evenly coated''
\item ``In a large bowl, mix sliced apples, sugar, cinnamon, allspice, clove and nutmeg''
\item ``In a large bowl, mix flour, baking powder, cinnamon, allspice, clove and nutmeg''
\end{enumerate}
Although (2) and (3) share more content,
(1) and (2) are semantically closer. The reason is that (1) and (2) correspond to the stuffing preparation phase, whereas (3) does not.

In particular, word embedding models (such as \cite{reimers2019sentence,cer2018universal,conneau2018senteval})
 are unlikely to capture a meaningful distance: in preliminary explorations we performed, those methods clustered together instructions with very different verbs and different ingredients.
 
 Thus, we decided to take advantage of the structure extracted by our parser and create a filtered list of \emph{candidate} pairs of instructions. We require that two instructions could be considered for the same cluster only if the verbs are similar \emph{and} they share enough ingredients, where ingredients that are common for the dish, such as apples, count more than rare ones (note that in the example above, (1) and (2) share more frequent ingredients, even though (2) and (3) share more ingredients in total). 
 {In the following, we explain the filtering steps and the clustering method.}
 


 \subsubsection{\textbf{Candidate pairs of instructions filtering}} \hfill \
 
 \xhdr{Verb similarity}
As word embedding models achieve poor performance on verbs \cite{schwartz2016symmetric},
we manually clustered the most frequent $100$ verbs in the data
and chose a representative verb per cluster. Then, we replaced verbs in the instructions with their representative.

\xhdr{Similarity of two ingredient objects}
To determine whether two ingredient objects are similar, we take into account their full ingredient names ($i_f^1, i_f^2$ correspondingly) and abbreviations ($i_a^1, i_a^2$), and check if:
\begin{displaymath}
max{\left(J\left(i_f^1,i_f^2\right),\ J\left(i_f^1,i_a^2\right),\ J\left(i_a^1,i_f^2\right),J\left(i_a^1,i_a^2\right)\right)}\geq t_1
\end{displaymath}
where $t_1\in[0,1]$ is a threshold and $J$ is the Jaccard index\footnote{$J(X,Y)=\frac{\left|X\cap Y\right|}{\left|X\cup Y\right|}$. In our case, $X,Y$ are the words in the ingredient name/abbreviation.}. For instance, for the name-abbreviation pairs: 
(grand smith apple, apple) and (red apple, apple) 
the similarity is 1. 

\xhdr{Similarity of two ingredient sets}
Let $I_1,I_2$ be two sets of ingredient objects; to measure their similarity, we use the weighted Jaccard similarity coefficient\footnote{$J_W(x,y)=\frac{\sum_j min(x_j,y_j)}{\sum_j max(x_j, y_j)}$ for two real vectors $x,y$.}  (also known as Ruzicka similarity), taking into account also the frequency of the items in $\mathcal{S}$. This coefficient can be restated as:
\begin{displaymath}
J_W\left(I_1,I_2\right):=\frac{\sum_{x\in I_1\cap I_2}\left(n_{x}\right)}{\sum_{y\in I_1\cup I_2}\left(n_{y}\right)}
\end{displaymath}
where $n_{x}$ is the number of recipes in which ingredient $x$ appears.

We consider $I_1,I_2$ to be similar if $J_W(I_1, I_2)>t_2$ for a threshold $t_2\in[0,1]$.
To calculate $J_W(I_1, I_2)$, the threshold $t_1$ must be set in advance, as computing ingredients' intersection or union expects knowing for every pair of ingredients $i_1\in I_1, i_2\in I_2$ whether they are similar or not.
We used Grid-Search on a dataset of 180 manually tagged ingredient list pairs to set values for these two hyperparameters (within bounds: 0-1), setting $t_1=0.35, t_2=0.325$.

{For instance, 
recall the example from the beginning of this section. 
The ingredients of the instructions (1), (2) and (3) are respectively: $I_1=$\{apples, sugar, cinnamon\}, $I_2=$\{apples, sugar, cinnamon, allspice, clove, nutmeg\},  $I_3=$\{flour, baking powder, cinnamon, clove, allspice, nutmeg\}.
Assuming apples appears 180 times in the multiple recipe set, flour 160 times, sugar 160, cinnamon 140, baking powder 90, nutmeg 35, clove 15, and allspice 10; then, $J_w(I_1,I_2)\approx 0.89$ and $J_W(I_2,I_3)\approx 0.25$.
That is, even though $I_2$ and $I_3$ share more ingredients in total, the similarity score of $I_1$ and $I_2$ is much higher as they share more \emph{frequent} ingredients.}



\subsubsection{\textbf{Training word2vec on recipes}} \label{word2vec_sec} \hfill \

After filtering pairs of instructions, taking advantage of the structured output of the parser, we can use a word embedding model to compute similarities. 
We trained a {CBOW variant} of  bigram word2vec model \cite{mikolov2013distributed} of dimension 100 on recipe instructions, {using \textit{Gensim} \cite{vrehuuvrek2011gensim}}. As mentioned in Section \ref{data}, in addition to the Allrecipes data, in this step we also included a large data set of recipes from ``Now You're Cooking!''.
Note that using the full instruction means that factors like instruments and time ranges are reflected in the embeddings.

\remove{
It is worth mentioning that we found it crucial that the sentence similarity model we use to compare two instructions would be trained specifically on recipes (see previous discussion in Section \ref{parser} \& \ref{clustering}).
Thus, we considered using BERT models for this task as they can be pretrained easily on specific data. 
However, acknowledging that using BERT for this task is time consumable and that the commonly used approach of averaging it output layer, yields bad sentence embeddings that are often worse than averaging GloVe embeddings \cite{pennington-etal-2014-glove, reimers2019sentence}, led us eventually choose in word2vec. 
}

\subsubsection{\textbf{The clustering method}} \hfill \


{We now define the similarity distance between two instructions that pass the filtering step (share a similar verb and enough ingredients) to be the cosine distance between their instruction embeddings (average of word embeddings). We define the distance between two instructions that do not pass the filtering step to be infinity.}

We chose hierarchical clustering with
complete-linkage criterion, merging clusters to the point when only infinitely distant clusters were left. We chose the linkage and stop criteria after
evaluating several criteria on three manually clustered dishes 
(that were not used for the evaluation).
Figure \ref{fig:main} shows a sample of instructions clustered together (to the right of the ``beat'' node).


%


{We note that we also experimented with clustering with constraints (e.g., forcing two instructions from the same recipe to be in separate clusters; taking into account the instructions' relative position in recipe). However, these approaches 
did not seem to improve the resulting clusters.}


%



\subsection{Constructing the summary graph}


We can now construct the summary graph $G=(V,E)$.
For every cluster, we define a corresponding vertex with weight equals to the number of instructions in it.
We also define source and target vertices \nstart\ and \nend.
 
We aim to connect vertices corresponding 
to subsequent actions. Hence,
for every two vertices
$v_l,v_k\in V\setminus \{\nstart,$ $\nend\}$, we consider $(v_l,v_k)\in E$ 
if there exist recipes in which an instruction from $v_k$ comes right after an instruction from $v_l$.
The edge weight $w(v_l,v_k)$ is  
the number of such recipes. Similarly, for every vertex $v\in V\setminus \{\nstart,\nend\}$ we consider $(\nstart, v)\in E$  (or $(v, \nend)\in E$) 
if there are instructions in $v$ that start (or end) a recipe.
The edge weight
is the number of such instructions.

\remove{
\begin{enumerate}
\item For every instruction cluster $S_i\in M$, we define a corresponding vertex $v_i$. In addition, we define two vertices $START$ and $END$ to be source and sink vertices correspondingly. Then, $V:= \{v_i\}_{i=1}^p\cup\{START, END\}$.
\item For every $v_l, v_k \in \{v_i\}_{i=1}^p$: $\left(v_l, v_k\right) \in E$ iff there is a recipe in which two \emph{subsequent instructions } $e_R^j, e_R^{j+1}$ are in $S_l$ and $S_k$ respectively. The edge weight $w(v_l, v_k)$ corresponds to the number of such recipes.
\item Similarly, for every $v_k \in \{v_i\}_{i=1}^p$: $(START, v_k) \in E$ 

(or $(v_k, END) \in E$) iff there is a recipe which its first (or last) instruction is in $S_k$. Again, the edge weight corresponds to the number of such recipes.


\end{enumerate}}


\xhdr{Pruning and noise reduction}
The graph is often too large to visualize effectively. Thus, we prune small clusters and weak edges, as well as nodes and edges that do not belong to a path from \nstart\ to \nend. We then choose up to 20 paths to be displayed to the user. 

Note that this pruning is only for visualization purposes, and the full graph is kept in memory. Pruned parts might be shown to the user as a part of the interaction (e.g., if they choose to explore a rare ingredient, light-weighted vertices and paths might be added back into the visualization). 

Ideally, we would have liked to display the 20 heaviest paths to the user.
However, picking out the heaviest simple paths in a graph is an NP-hard problem. Thus, we resorted to a heuristic {approach} adapted from \cite{filippova2010multi}.
%
{First, we invert the edge weights and search for K-shortest paths in terms of the edge weights (with a K that is sufficiently bigger than the number of 
paths we finally display to the user).\footnote{As we wished to display 20 paths to the user, we set K to 60.} 
%
As our edges are added locally, some short paths do not actually represent a full recipe (e.g., if there are parts of the recipe that could be carried out in a different order, this creates a cycle in the graph that can be shortcut). Thus, we filter out paths that are too short (number of edges). This bound is set to be the minimal recipes' number of instructions after trimming 10\% of the smallest values. 
Finally, we rerank the remaining paths by normalizing their weights over their lengths. The highest 20 ranked paths are chosen to compose the graph displayed to the user.}

Importantly, noisy instructions are likely to either become small clusters and be pruned or join an existing, large cluster and have virtually no effect on its summary (what the user sees). 

Building the graph for $\sim200$ parsed recipes takes around 1-2 minutes on a personal computer.



\xhdr{Visualization}
We built a user interface using React.js showing the {compact version of the} summary graph (refer again to Figure \ref{fig:main}). Dark nodes contain more instructions, and thick edges represent strong connections between the nodes. Every cluster is represented by the main verb and  a summarization of the ingredients. Ingredients are accompanied with relative frequency in the cluster; clicking on node reveals quantity range (normalized to the most frequent number of servings), tools and time range. The user can also choose to see the full list of instructions. Further actions include seeing lists of common and rare ingredients, tracking ingredients through a graph, and multi-faceted filtering. User interactions (such as requesting specific ingredients) might result in uncovering paths that were hidden before, as they were not in the {20 chosen paths.}
%


\section{Evaluation}
We now turn to evaluating our representation. We wished to answer three main questions: (1) Is the representation {\bf intuitive}, (2) Is the representation {\bf coherent} (i.e., do paths correspond to recipes), and (3) Is the representation {\bf useful}.
We recruited in total 50 participants, including 10 experts. 
Following the recommendation of \cite{nielsen1993mathematical}, we chose to run multiple tests with 11-20 users in each. This also had the benefit of being able to closely \emph{observe} all the participants using the system.


We randomly sampled a set of dishes from the most popular categories (soup, salad, cake etc.), having at least 100 recipes each. For the experiments we sampled four dishes out of this set: two simple ones (guacamole, omelette) and two complex ones (apple cake, spaghetti), judged by the average number of instructions.

\subsection{Intuitiveness and coherence of representation} \label{clarity_eval}

We started by evaluating the intuitiveness and coherence of the summary graph. 20 student volunteers were recruited to this experiment. 11 participated in part I, and all participated in part II.

\xhdr{Part I: Intuitiveness}
We showed participants the UI for one of the four dishes without providing any explanation.
We asked them to explain what nodes, edges and paths  from \nstart\ to \nend\ represent.
Full, accurate response rates were $81.8\%, 90.9\%, 90.9\%$ (nodes, edges, paths respectively). 
The others provided partially correct explanations. 
For example, one participant wrote that an edge represents ``a transition between steps in the preparation of the recipe", and a path represents ``all steps in a recipe", but described a node as ``the most common ingredients", which we considered too vague.
Thus, we conclude that the graph is indeed mostly intuitive.

\xhdr{Part II: Coherence}
In this part we provided the participants with a brief explanation of the UI. 
Our goal was to test the coherence of the representation (Do nodes correspond to instructions? Do paths correspond to recipes?), once the participants understood the UI.

\xhdr{\textit{\textbf{Recipes and paths}}}
 Given a random recipe, we asked the participants to mark a path on the graph that fits it best, achieving $75\%$ success rate. 
 Afterwards, we asked for the opposite, writing a recipe corresponding to a random marked path, achieving $90\%$ success. 
 
{When translating recipes to graphs, most failures were a result of the participants searching for a path that fitted the given recipe \textit{exactly} (although they were told to mark the one that {fits it best}).
In translating graphs to recipes, one non-native English speaker participant failed to understand some of the instructions; another wrote ``go over the nodes and perform the steps described in them'' but did not provide an explicit recipe. 
}
These results suggest it is relatively straightforward to translate between recipes and paths. 

\xhdr{\textit{\textbf{Node and path coherence}}}
We asked the participants to: (1) pick three nodes and rate the coherence of the instructions within them;
(2) follow two random {marked} paths and rate how much they represent a possible recipe. 
We also asked them to (3) rate the graph according to its comprehensibility.

All ratings were in a Likert scale of 1-5 \cite{likert1932technique}. See Table~\ref{tab:table1} for results. Results are encouraging overall, with nodes rated as very coherent, paths rated as good, and the graph as comprehensible.

\begin{table}[t!]
\centering
\scalebox{0.87}{%
\begin{tabular}{lccc}
\hline \textbf{Measures} & \textbf{\# of scores} & \textbf{Average}  & \textbf{Std}
\\ \hline
(1) Node coherence & 60  & 4.55  & 0.723  \\
(2) Reasonable paths & 40  & 3.825  & 1.196  \\
(3) Graph comprehensibility (1st exp.) & 20  & 4  & 0.973  \\
(3*) Graph comprehensibility (2nd exp.) & 20  & 3.85  & 0.72  \\
\hline
\end{tabular}
}
\caption{\label{tab:table1} 
Graph's clarity \& coherence (Likert scale, 1-5). 
}
\end{table}

\subsection{Utility to users}
After evaluating the representation, we turned to evaluate its utility to users. We recruited another 20 student volunteers who were randomly divided into two groups (A and B), and asked them to rate their cooking level of expertise (Likert scale, 1-5. \textit{group A:} mean level score was 3.5, std=0.92. \textit{group B:} mean=3.6, std=0.8).

To the best of our knowledge, no benchmark exists for graphs that summarize many procedural texts. Thus, as a baseline we simulated what people are likely to use today -- recipe books and websites. In this condition, users received a searchable file that contained hundreds of recipes {for the same dish}.

We focused on two dishes: guacamole (easy) and apple cake (harder). Each user saw both dishes: Group A worked on the guacamole dish first and group B on the apple cake first. For the first dish, users received a searchable file (similar to cooking books and recipe websites); in the second, they got the UI. 
For both dishes, we asked the users to perform the following two tasks:

\begin{enumerate}
\item \textbf{Clarifying} a recipe for a novice cook. We asked participants to add missing details that might not be trivial to a novice cook (e.g., add an important action such as cool-after-baking; explain vague descriptions such as ``al-dente'', or  specify an exact amount of salt instead of ``to taste''), and replace unusual things with more common ones.


\item \textbf{Adding a creative twist} to spice up a recipe.
\end{enumerate}
We limited the time for completing each task to 8 minutes. In the clarifying task, to avoid the case participants know the answer from experience, we required references supporting their answers.
After each task, the participants rated how hard it was for them, and were encouraged to share insights about the dish.
At the end, they were asked to provide general feedback and, as in the experiment of Section \ref{clarity_eval}, to rank the graph's comprehensibility. 

For the first task (clarifying), we selected recipes with 
common mistakes.\footnote{\href{https://tinyurl.com/guacamole-common-mistakes}{\color{blue(pigment)} https://tinyurl.com/guacamole-common-mistakes}, \\ \href{https://tinyurl.com/cake-common-mistakes}{\color{blue(pigment)} https://tinyurl.com/cake-common-mistakes}.}
For the second (adding a twist)
we picked the simplest recipe in the data, in terms of number of actions and ingredients.

\xhdr{Participants' insights and feedback}
We were encouraged to find that participants identified almost twice more insights when using the graph (13 vs.~7). 
Overall, feedback was very positive. Snippets include: 
``It was such a relief using the summary graph after having to go over so many recipes'', 
``This graph is awesome!'',  
``The statistics information was very handy and accessible. I wish all my recipes were shown to me in such form''. Negative feedback focused mostly on the UI, and not on the content of the graph itself.
%
%


\xhdr{A note on fixation}
While observing participants performing
the tasks, we noticed that many baseline users fixated on one recipe (often the first one on the list). As one user
explained in their feedback, ``I decided to focus on one recipe
and base most of my modifications on it. The graph gave
a more global view from which I could infer changes more
easily''.

\xhdr{Participants' output}

\paragraph{\textit{\textbf{Verifying feasible outcomes:}}}
 Since our edge creation method is local, we wanted to verify 
 that the usage of the graph can still yield feasible outcomes.
Thus, we asked three cooking experts to rate the feasibility of all the experiment's outcomes
on a Likert scale of 1-7. We measured the mean score, resulting in 5.65 for the file (std = 1.515) and 5.63 using the graph (std = 1.461). Thus, we conclude that using the graph does not change the feasibility of the users' outcome. 

\begin{figure} \label{img:adjusting}
\centering
\includegraphics[width=1\linewidth]{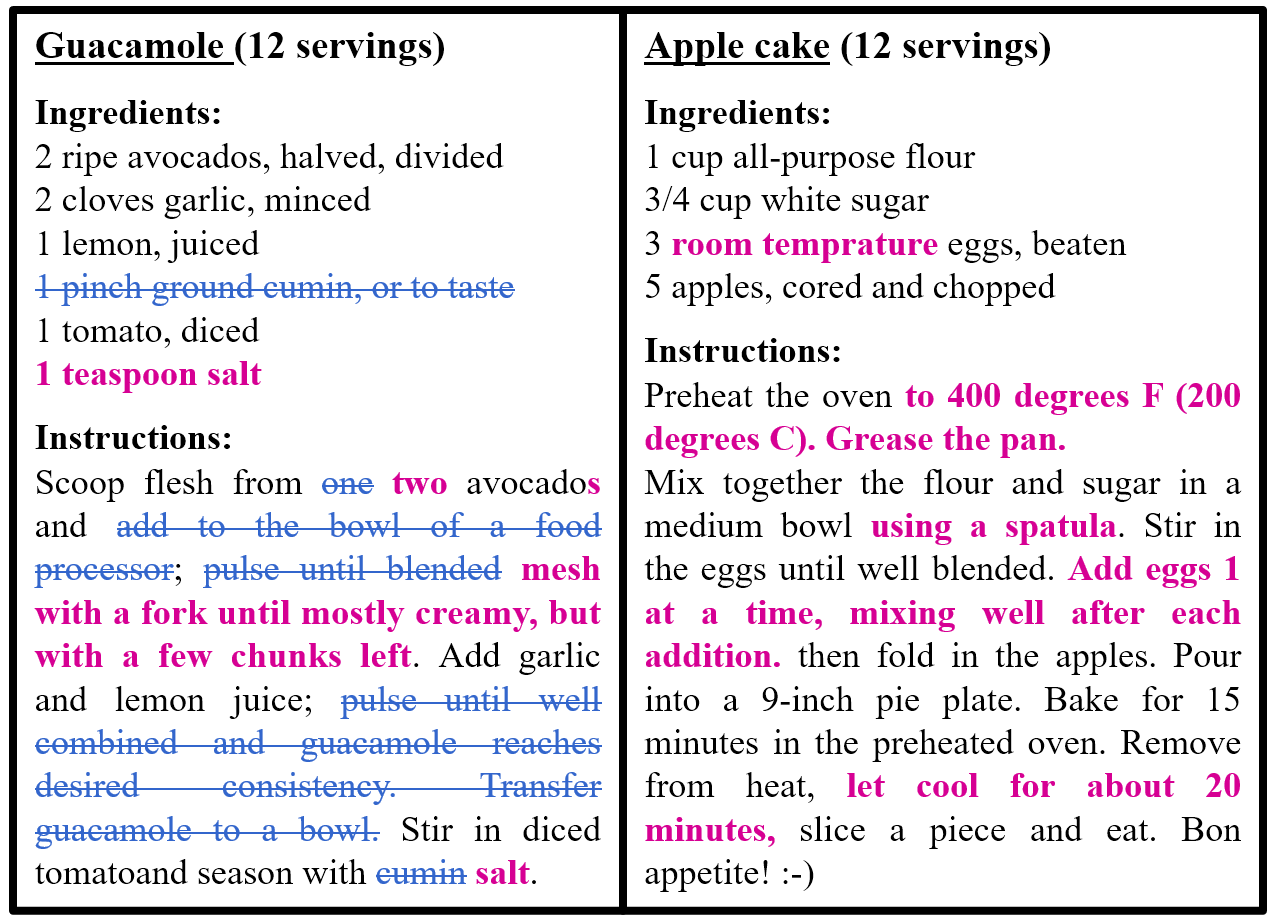}
\caption{\label{fig:adjustments} 
Two of the participants' modified recipes after clarifying them for the novice cook with the support of the graph. For example, in the left modified recipe (guacamole), the participant decided to mash the avocado with a fork instead of food processor. In the right modified recipe (apple-cake), the participant realized that the greasing-the-pan action was missing and added it. 
}
\end{figure}

\paragraph{\textit{\textbf{Clarifying for the novice cook:}}}
{After collecting all the changes suggested by the participants (see Figure \ref{fig:adjustments} for examples of adjusted recipes),}
we recruited two cooking expert to annotate whether changes suggested by participants: (1) could really assist a novice cook, (2) could be crucial for the recipe to succeed.

Our experts had good agreement --
for guacamole we measured Cohen's Kappa=0.661 \cite{cohen1960coefficient}, accuracy=0.867; for apple cake Kappa=0.593 and accuracy=0.806. 
We took only changes chosen by both annotators as ground truth and counted how many
were detected by each participant. For the more complex dish (apple cake), participants performed significantly better using the graph (the average number of changes without the graph was 1.9, with the graph 3.7, p-value = 1.06E-05; \textit{critical changes}: 1.4 without the graph, 3.1 with, p-value = 7.68E-07; independent samples t-test). 
For the simpler dish (guacamole) there was only a slight advantage in favor of the graph. These results are compatible with our intuition that the graph can help more with complex recipes.

We also tested whether the more experienced cooks (10 people; cooking expertise 4-5), being more aware of nuances, performed significantly better using the graph. It was indeed the case for both dishes (2.5 on average without the graph, 3.9 with, p-value = 0.014; \textit{critical changes:} 1.1 without, 2.4 with, p-value = 0.0016; independent samples t-test). 


\remove{
\begin{table}[t!]
 \centering
 \scalebox{1.05}{%
 \begin{tabular}{lccc}
 \hline \textbf{Task} &  \textbf{Supported by graph?} & \textbf{Average}  & \textbf{Std} \\ \hline
\textbf{Clarifying}  & X  & 3.3 & 1.1   \\
 & V  & 2.8 & 0.75  \\
\textbf{Adding a twist}  & X  & 2.2 & 1.17 \\
 & V  & 2.1 & 1.22  \\
\hline
\end{tabular}}
\caption{\label{tab:table2} 
Gaucamole: difficulty level statistics per task (with and without the graph support).
}

 \centering
 \scalebox{1.05}{%
 \begin{tabular}{lccc}
 \hline \textbf{Task} &  \textbf{Supported by graph?} & \textbf{Average}  & \textbf{Std}\\ 
 \hline
\textbf{Clarifying}  & X  & 3.3 & 0.9  \\
& V  & 2.5 & 0.67  \\
\textbf{Adding a twist} & X  & 2.4 & 1.11 \\
& V  & 1.6 & 0.8 \\
\hline
\end{tabular}}
\caption{\label{tab:table3} 
Apple-cake: difficulty level statistics per task (with and without the graph support).
}
\end{table}
}

\begin{figure}
\centering
\includegraphics[width=\linewidth]{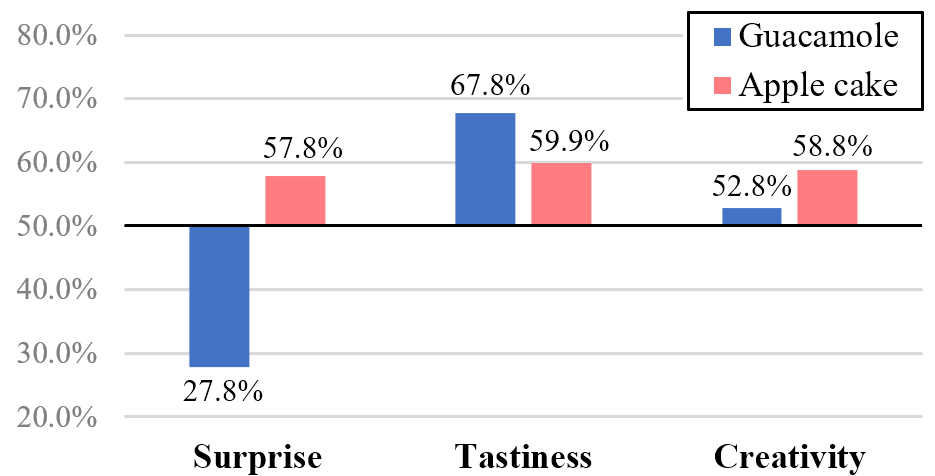}
\caption{\label{fig:cretivity_graph} 
Percentages of times that
unique graph's ingredients
beat unique list's ingredients. 
Comparisons are computed within participant.
}
\end{figure}

\paragraph{\textit{\textbf{Adding a twist:}}}
To reduce individual bias, we collected the two groups' unique ingredients (i.e., those appeared in one group and not in other) and compared them. For that, we asked five cooking experts to rank (Likert scale, 1-5) each ingredient in terms of: (1) \textbf{surprise} (how surprising it is for the dish?) and (2) \textbf{tastiness} (how suitable it is in terms of taste?). We then computed for every pair of ratings a \textbf{creativity} score, which we defined as the \emph{minimum} of these ratings. Creativity is often defined as a combination of novelty and value \cite{gaut2010philosophy,lamb2018evaluating};
We chose the minimum since we wanted this score to reflect both the novelty (surprise) and the value (tastiness).

Likert scores are difficult to compare among different people. Thus, for each expert, we made pairwise comparisons between each two ingredients they rated, and computed the percentages of times an ingredient from one group beats ingredients from the other. 

\begin{figure}
\centering
\includegraphics[width=\linewidth, trim={0 0 0 0cm}, clip]{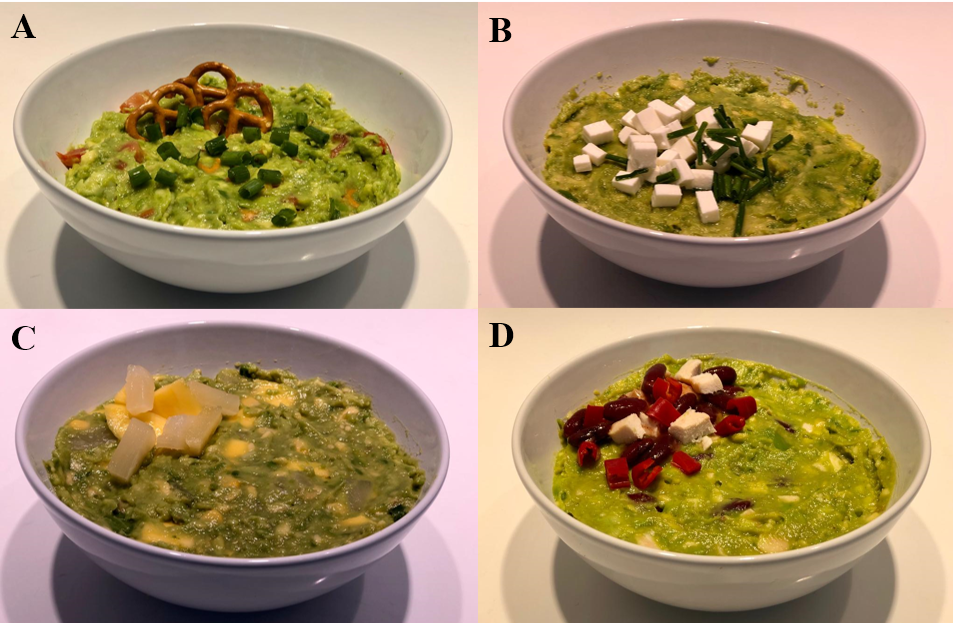}
\caption{\label{fig:fig7} 
Examples of participants' creative guacamole dishes obtained with the support of the graph: (A) guacamole with pretzel fragments, cherry tomatoes and green onion; (B) guacamole with feta cheese and chives; (C) guacamole with mango, pineapple and corn; (D) guacamole with chicken breast, red beans and jalapeno.
}
\end{figure}

The results are in Figure \ref{fig:cretivity_graph}. 
For the apple cake dish, the graph's ingredients beat those of the file in all parameters.
For the guacamole dish, graph ingredients won in terms of tastiness and creativity but not surprise.
Looking closer at the results, we observed that baseline users often made ingredients up (not basing them on a recipe), while graph users observed ingredients used in recipes and tried to generalize them (e.g., different salty snacks or different tropical fruit), which might explain these findings.
Tables \ref{tab:gua_c}  and \ref{tab:apple_c} show the winning unique ingredients in terms of creativity and their origin (graph or baseline). Figure \ref{fig:fig7} shows a sample of four prepared guacamole dishes based on the graph users suggestions.

\xhdr{A note on task difficulty} 
After each task, the participants 
rated its difficulty on a scale of 1-5 (1 stood for ``piece of cake'',\footnote{No pun intended} and 5 for ``extremely difficult''). 
Results are in Table \ref{tab:table2}.
Overall, the tasks where the user had access to the graph were rated as easier than those supported by the list (baseline),
 but the effect was not large. 
The change was most pronounced in the clarifying
and creativity tasks for the more complex dish (both statistically significant, p-values = 0.023, 0.048).

\begin{table}[t!]
\centering
\resizebox{\columnwidth}{56pt}{
\begin{tabular}{clcc}
\hline \textbf{Rank} & \textbf{Ingredient} & \textbf{Total creativity score} & \textbf{Origin}
\\ \hline
1 & pretzel fragments & 18 & graph \\
2 & parmesan cheese & 17 & graph \\
3 & barbeque pringles & 16 & list \\
- & kidney bean & 16 & graph \\
5 & soup nuts & 15 & graph \\
- & chicken breast & 15 & graph \\
- & cream cheese & 15 & graph \\
8 & sour cream & 14 & list \\
- & balsamic vinegar & 14 & graph \\
10 & bulgarian cheese & 13 & list \\
\hline
\end{tabular}
}
\bigskip
\caption{\label{tab:gua_c} 
The top-ten ranked ingredients in terms of creativity for the guacamole dish. There were in total 25 ingredients to compare after collecting the two group's unique ingredients, 9 came from the list and 16 from the graph.
}

\centering
\resizebox{\columnwidth}{56pt}{
\begin{tabular}{clcc}
\hline \textbf{Rank} & \textbf{Ingredient} & \textbf{Total creativity score} & \textbf{Origin}
\\ \hline
1 & blueberries & 19 & graph \\
- & cherries & 19 & graph \\
3 & cranberry juice & 17 & graph \\
- & wrapped caramels & 17 & list \\
5 & coconut flour & 15 & graph \\
6 & candied lemon & 14 & list \\
- & shredded coconut & 14 & graph \\
- & banana & 14 & graph \\
9 & lotus spread & 13 & graph \\
- & carrots & 13 & graph \\
\hline
\end{tabular}
}
\bigskip
\caption{\label{tab:apple_c} 
The top-ten ranked ingredients in terms of creativity for the apple-cake dish. There were in total 34 ingredients to compare after collecting the two group's unique ingredients, 19 came from the list and 15 from the graph.
}
\end{table}

\begin{table}[h!]

 \centering
 \resizebox{\columnwidth}{26.5pt}{
 \begin{tabular}{lc|cc|cc}
 \hline \textbf{Task} &  \textbf{Graph support?} & \textbf{Avg.}  & \textbf{Std} & \textbf{Avg.} &  \textbf{Std}\\ \hline
\textbf{Clarifying}  & X  & 3.3 & 1.1 & 3.3 & 0.9  \\
 & V  & 2.8 & 0.75 & 2.5 & 0.67 \\
\textbf{Adding a twist}  & X  & 2.2 & 1.17 & 2.4 & 1.11\\
 & V  & 2.1 & 1.22 & 1.6 & 0.8 \\
\hline
\end{tabular}
}
\bigskip
\caption{\label{tab:table2} 
Difficulty level statistics per task with and without the graph support (left: guacamole, right: apple cake).}
\end{table}

\remove{
\xhdr{A note on fixation}
While observing participants performing the tasks, we noticed that participants viewing the file tended to \emph{fixate} on one recipe (often the first one on it). As one user explained in their feedback, “I decided to focus on one recipe and base most of my modifications on it. The graph gave a more global view from which I could infer changes more easily”. }

\smallskip

While preliminary, we believe these studies demonstrate the potential of the summary graph representation in helping people navigate (and make sense of) a large body of procedural texts.

\section{Related work} 
Our work is related to multiple lines of work.

\xhdr{Sensemaking}
Broadly speaking, the goals of our system align with these of the \emph{sensemaking} domain.
As described by D. M. Russel \shortcite{russell1993cost}, \emph{Sensemaking} is the task of constructing a mental representation of
interrelated pieces of information relevant to 
answering task-specific questions, often in the context of understanding large document collections.
In this paper, the interrelated pieces are a large collection of procedural texts sharing the same goal, and the aim is to help users understand them and easily explore commonalities and differences in them.
Sensemaking has been studied extensively in various fields, including 
HCI \cite{russell1993cost, hahn2016knowledge}, 
information science \cite{dervin2003human, havn2006sensemaking, griffith1999technology}, organizational science \cite{weick1995sensemaking} and education \cite{arcavi1992mathematics}. 
As opposed to our work, much of the sensemaking work relies on crowdsourcing for aggregating and arranging the different pieces of information. 

{
\xhdr{Multi-document summarization}
Our work is also related to  multi-document summarization, }
{and in particular to graph-based multi-document summarization approaches \cite{zhao2020summpip,  giannakopoulos2014newsum, baralis2013graphsum, filippova2010multi, erkan2004lexrank, mihalcea2004textrank}.
These works also represent document units as graphs, on which they apply graph-based ranking algorithms to generate a summary. However, the output is a text (the summary) and not a graph that allows users to explore commonalities and differences between the texts. We are also not aware of such methods applied to procedural texts. 
}

\xhdr{Procedural texts}
Understanding procedural texts is the base of a substantial body of research within natural language understanding \cite{malmaud2014cooking,tasse2008sour,delpech2008investigating,delpech2007two,bollini2013interpreting,beetz2011robotic,misra2016tell, shridhar2020alfred}. 
A prominent line of work suggests ways to transform natural language instructions into a graph structured representation
\cite{karikome2018flow,chen2011learning,patow2010user,yamakata2016method,kiddon2015mise,mori2014flow,maeta2015framework}. 
{These works use a graph to represent a \emph{single} procedural text. In contrast, we summarize \emph{many} procedural texts into a single graph.
We believe our representation could aid users in performing sensemaking tasks such as modifying a given procedure to satisfy individual preferences or constraints.}


\xhdr{Cooking Recipes}
Much research regarding procedural texts focuses on cooking recipes \cite{mujtaba2020towards,li2020recipes,min2019survey}. 
Importantly, most work does not make changes to recipes, but instead focuses primarily on recommending recipes from an existing pool
\cite{theodoridis2019survey,elsweiler2017exploiting,ueda2014recipe,freyne2010intelligent,forbes2011content,trattner2017food,teng2012recipe}.
Recently, Majumder et al.~\shortcite{majumder2019generating} sought to combine work from recommender systems and text generation. 
However, their system gives the user only a little control over the text being produced.

Perhaps the closest work to ours is a work by Chang et al.~\shortcite{chang2018recipescape}, which assists cooking experts and culinary students in browsing and comparing hundreds of recipes via an interactive system. However, their use case is very different from ours, as their output summarizes only some of the aspects of recipes, providing a very different view of the landscape, meant for an audience of experts.

We also note that the idea of aggregating recipes has been suggested before, sometimes jokingly, in popular culture. The book ``Cooking for Geeks'' \cite{potter2010cooking} includes a recipe for the ``Average Internet Pancakes'', noting that {\it ``No one’s ever wrong on the Internet, so the average of a whole bunch of right things must be righter, right? The quantities here are based on the average of the eight different pancake recipes from an online search''}. The website ThePudding took this idea one step further, taking 200 chocolate chip cookie recipes and trying to generate the average cookie using a mathematical average, predictive text algorithms, and neural networks.\footnote{\href{https://pudding.cool/2018/05/cookies/}{\color{blue(pigment)} https://pudding.cool/2018/05/cookies/}}


\section{Conclusion and future work}

The web is full of procedural texts, many of them sharing the same goal.
When performing sensemaking tasks one needs to be able to view the bigger picture; however, this is often time-consuming, requiring extensive browsing and comparisons. 

In this work we proposed {a novel} unsupervised learning approach for which the input is a set of procedural texts sharing the same goal, and the output is an intuitive graph representation summarizing them, mapping the landscape of possibilities. 
We believe this representation could allow users to explore commonalities and differences between the various ways to carry out a task and devise a way to accomplish the task.

{We demonstrated our system on \emph{cooking recipes}, a promiment example of procedural texts.
We devised an unsupervised recipe parser, taking into account the unique structure of recipes, and proposed an algorithm for constructing the summarization graph.
User studies showed that our representation is easy to work with and could help users with several sensemaking tasks, such as understanding or modifying a recipe.}

In the future, we plan to apply the proposed approach to other domains. For example, many scientific areas use procedural texts (material science, manufacturing medicine). Using a graph representation might help the scientist gain knowledge and insights into the process.
Another exciting avenue is exploring the creativity-supporting aspects of the graph. We believe identifying anomalies in the graph could help surfacing creative options. 

Beyond the specific application in this paper, we envision a future where fully automated systems can digest a large set of procedural texts, answering queries and modifying the texts according to user needs and preferences.

\section*{Acknowledgements}
We thank the anonymous reviewers for their insightful comments, Hyadata Lab members for thoughtful remarks, and the participants in our user studies.
This work was supported by the European Research Council (ERC) under the European Union's Horizon 2020 research and innovation programme (grant no. 852686, SIAM).

\pagebreak
\bibliographystyle{ACM-Reference-Format}
\bibliography{__bibliography}


\end{document}